\documentclass[10pt,twocolumn,letterpaper]{article}

\usepackage{iccv}
\usepackage{times}
\usepackage{epsfig}
\usepackage{graphicx}
\usepackage{amsmath}
\usepackage{amssymb}
\usepackage{sidecap}
\usepackage{xfrac}
\usepackage{arydshln}
\usepackage[table, dvipsnames]{xcolor}
\usepackage[]{algorithm2e}
\usepackage[accsupp]{axessibility} 

\usepackage{multirow}
\usepackage{tabularx} 
\newcommand\setrow[1]{\gdef\rowmac{#1}#1\ignorespaces}
\newcommand\clearrow{\global\let\rowmac\relax}
\newcommand{\mc}{\mathcal}
\usepackage{adjustbox}
\usepackage{url}
\usepackage{floatrow}
\usepackage{float}
\usepackage{subfig}
\newfloatcommand{capbtabbox}{table}[][\FBwidth]
\SetKwInput{KwOutput}{Output}              
\newlength\mylen

\newcommand\largestbox[1]{\textcolor{black}{#1}}
\newcolumntype{a}{>{\columncolor{gray!30}}c}

\usepackage{breakurl}

\usepackage[pagebackref=true,breaklinks=true,letterpaper=true,colorlinks,bookmarks=false]{hyperref}
\usepackage{cleveref}
\iccvfinalcopy 

\def\iccvPaperID{8247} 

\begin{document}

\pagenumbering{gobble}


\title{Towards Rotation Invariance in Object Detection}

\author{Agastya Kalra$^{1}$, Guy Stoppi$^{1}$, Bradley Brown$^{1}$, Rishav Agarwal$^{1}$ and Achuta Kadambi$^{1,2}$\\
$^{1}$Akasha Imaging, Palo Alto CA\\
$^{2}$UCLA, Los Angeles CA\\
{\tt\small \{agastya, guy.stoppi, bradley.brown, rishav\}@akasha.im, \{achuta\}@ee.ucla.edu}
}
\maketitle


\begin{abstract}
Rotation augmentations generally improve a model's invariance/equivariance to rotation - except in object detection. In object detection the shape is not known, therefore rotation creates a label ambiguity. We show that the de-facto method for bounding box label rotation, the Largest Box Method, creates very large labels, leading to poor performance and in many cases worse performance than using no rotation at all. We propose a new method of rotation augmentation that can be implemented in a few lines of code. First, we create a differentiable approximation of label accuracy and show that axis-aligning the bounding box around an ellipse is optimal. We then introduce Rotation Uncertainty (RU) Loss, allowing the model to adapt to the uncertainty of the labels. On five different datasets (including COCO, PascalVOC, and Transparent Object Bin Picking), this approach improves the rotational invariance of both one-stage and two-stage architectures when measured with AP, AP50, and AP75. The code is available at \url{https://github.com/akasha-imaging/ICCV2021}.
\end{abstract}


\section{Introduction}

It is desirable for object detectors to work when scenes are rotated. But there is a problem: methods like Convolutional Neural Networks (CNNs) may be scale and translation invariant but CNNs are not rotation invariant~\cite{Goodfellow-et-al-2016}. To overcome this problem, the training dataset can be expanded to include data at new rotation angles. This is known as \emph{rotation augmentation}. In object detection, rotation augmentation can be abstracted as follows: given an original bounding box and any desired angle of rotation, how should we determine the axis-aligned rotated bounding box label? If the shape of the object is known, this is quite simple: we rotate the object shape and re-compute the bounding box. However in the case of object detection, the shape is unknown. 

The prevailing folk wisdom in the community is to select a label large enough to completely overlap the rotated box~\cite{zoph2019learning}. In studying this problem, we find that this method may hurt performance, and on COCO~\cite{lin2014microsoft}, we find that \emph{every other prior we tried is better}. Yet somehow this \textbf{``Largest Box"} method is very prevalent both in academia and in large scale object detection codebases~\cite{xi2018sr, bochkovskiy2020yolov4, zoph2019learning, tan2019efficientnet,montserrat2017training, liu2016novel, albumentations-team, chen2019mmdetection, abadi2016tensorflow, jung2018imgaug,casado2019clodsa, solt2019}. Indeed, recent analysis has found that largest box is only robust to about $< 3^\circ$~\cite{fastai} of rotation.


\begin{figure}
\begin{center}
    \centering
    \includegraphics[width=\textwidth]{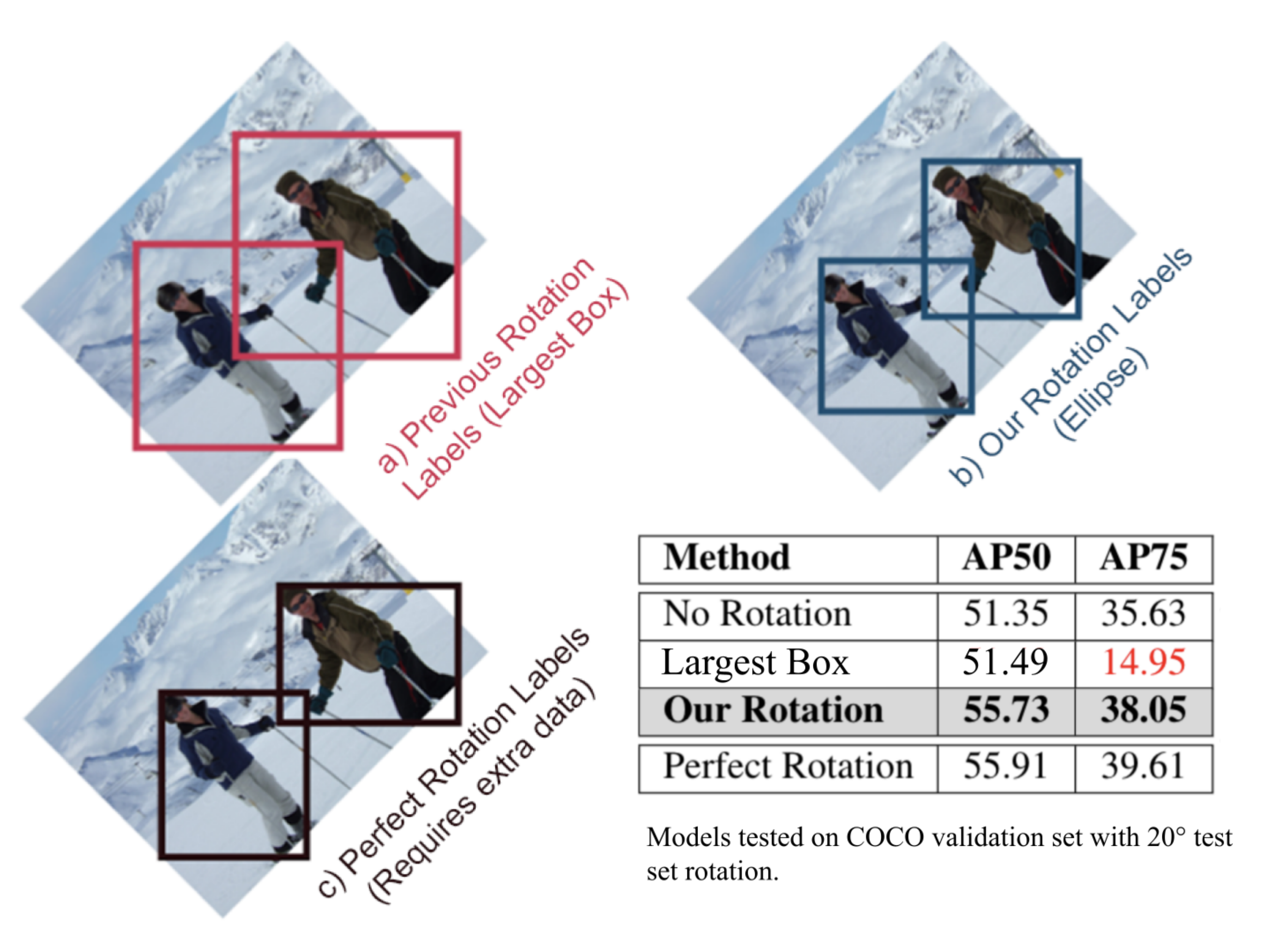}
    \caption{\textbf{A method is proposed to properly rotate a bounding box for rotation augmentation.} The previous solution of largest box is an overestimate of the perfect bounding box for a rotated scene. See table for how choice of rotation augmentation affects object detection performance. \label{fig:first}}
\end{center}%
\end{figure}

In this paper, we propose an advance on the largest box solution that achieves significantly better performance on five object detection datasets; while retaining the simplicity of a few lines of code implementation. 

\subsection{Contributions}

In a nutshell, our solution has two aspects. First, we derive an \textbf{elliptical shape} prior from first principles to determine the rotated box label. We compare it to many other novel priors and show this is optimal. Second, we introduce a novel \textbf{Rotation Uncertainty (RU) Loss} function, which allows the network to adapt the labels at higher rotations using priors from lower rotations based on label certainty. We demonstrate the effectiveness of this solution by both improving performance datasets where rotation is important such as Pascal VOC~\cite{everingham2010pascal} and Transparent Object Bin Picking~\cite{kalra_2020_cvpr} and generalizing to novel test time rotations on MS COCO \cite{lin2014microsoft} (Figure~\ref{fig:first}).

\subsection{Scope} Rotation data augmentation in object detection is not new. This paper is not about finding the best overall way to use a rotation data augmentation. For that - a brute force search or papers like AutoAugment~\cite{zoph2019learning} might be better examples. This paper focuses solely on methods to perform a rotation augmentation on axis-aligned bounding boxes. When implemented, these proposed modifications boil down to a few lines of code and leave little reason to use the current Largest Box method. 

\section{Related Work}

\textbf{Data Augmentations} are an effective technique to boost the performance of object detection. Data augmentation increases the quantity and improves the diversity of data. Data augmentations are of two types. Photo-metric transforms modify the color channels such that the detector becomes invariant to change in lighting and color. Classical photometric techniques include adding Gaussian blur or adding colour jitters. Modern photo-metric augmentations like Cutout~\cite{devries2017improved} and CutMix~\cite{yun2019cutmix}, randomly remove patches of the image. On the other hand, geometric transforms modify the image's geometry, making the detector invariant to position and orientation. Geometric modifications require corresponding changes to the labels as well. Geometric transforms are  difficult to implement ~\cite{albumentations-team} and contribute more towards accuracy improvements~\cite{taylor2018improving}. We focus specifically on rotation augmentations and object detection.   

\begin{figure}
\begin{center}
    \centering
    \includegraphics[width=\textwidth]{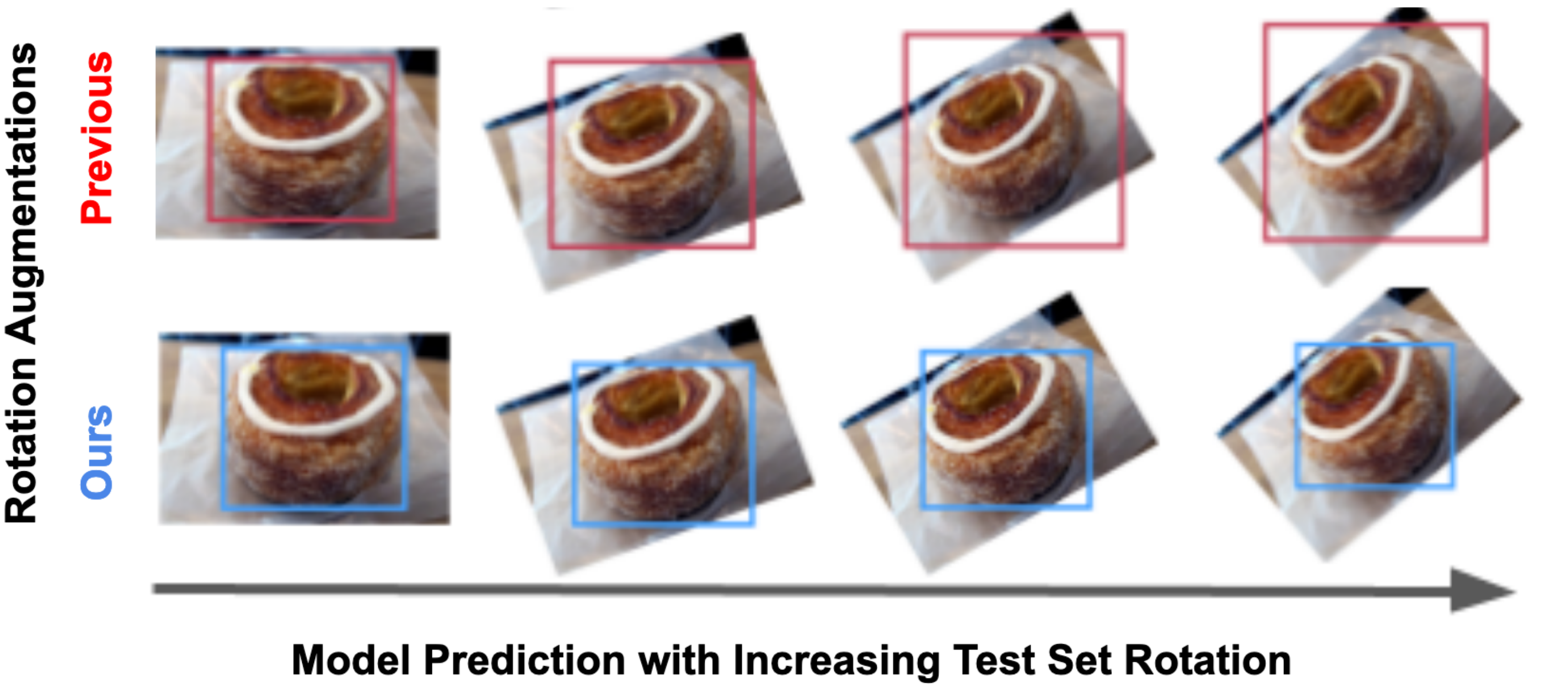}
    \caption{\textbf{Comparison between our method and other methods.} We show example predictions from models trained with each rotation augmentation above. Our method has comparable performance to using perfect segmentation labels without requiring the extra shape information. \label{fig:method_compare}}
    \vspace{4mm}
    
        \resizebox{\linewidth}{!}{
\begin{tabular}{|l | c | c | c | c | c}
\hline
  \textbf{ Rotation Method} & \textbf{AP50 (Coarse)} & \textbf{AP75 (Fine)}  &\textbf{ Shape Label} \\\hline
    No Rotation  & \cellcolor{orange!25}Med & \cellcolor{orange!25}Med &  \cellcolor{green!25}No\\
  \hline
    Largest Box (e.g.~\cite{abadi2016tensorflow, jung2018imgaug, albumentations-team})  & \cellcolor{orange!25}Med & \cellcolor{red!25}Low &  \cellcolor{green!25}No\\
  \hline
 \setrow{\bfseries} Ellipse + RU Loss
(Ours) & \cellcolor{green!25}{Very High} & \cellcolor{green!25}{High} & \cellcolor{green!25}{No}\\
\hline
  Perfect Box (Gold Std) & \cellcolor{green!25}Very High & \cellcolor{green!25}Very High & \cellcolor{red!25}Yes\\
\hline
\end{tabular}
}
\end{center}

\end{figure}

\textbf{Rotation Augmentation in Object Detection} is currently done by the largest box method for major repositories (e.g.~\cite{albumentations-team, chen2019mmdetection, abadi2016tensorflow, jung2018imgaug,casado2019clodsa, solt2019}) and publications (e.g.~\cite{xi2018sr, bochkovskiy2020yolov4, zoph2019learning, tan2019efficientnet,montserrat2017training, liu2016novel}) that do bounding box rotation for deep learning object detection. The largest box method does a great job guaranteeing containment, but at large angles, it severely overestimates the bounding box's size. Figure \ref{fig:method_compare} shows an example of these over-sized bounding boxes. For that reason, FastAI ~\cite{fastai} recommends rotation of no more than 3 degrees. Some recent work, such as AutoAugment \cite{zoph2019learning, tan2019efficientnet}, use rotation as part of a complex learned data augmentation scheme. While learning rotation augmentation directly is interesting, it requires extensive computing resources. We seek to achieve the simplicity of the largest box, with performance improvements for larger angles. 

\textbf{Oriented Bounding Boxes}, a sister of object detection, is the task of predicting non-axis aligned bounding boxes, also known as oriented bounding boxes. Several methods like~\cite{Ding_2019_CVPR, cheng2018learning, liu2017learning}, aim to achieve rotation invariance when predicting rotated bounding boxes. However, these methods already have labelled rotated boxes as input and do not end up with loose boxes when the input image is rotated.  As this is a different task, it is out of the scope of this paper. Our paper focuses on axis-aligned bounding boxes only.

\textbf{Rotational Invariance} is an important problem to solve in object detection. Classical computer vision methods achieved rotational invariance by extracting features from images~\cite{wang2015ordinal, liu2018rotation,liu2014rotation,schmidt2012learning}. With the rise of neural networks, newer methods attempt to modify the architecture to achieve rotation invariance ~\cite{cheng2016rifd,cheng2016learning,xi2018sr}. These methods rotate the input images and add special layers that learn the object's orientation in the image. Our general-purpose method can again aid these methods as well.

\section{Background}
\label{s:bgnd}

\subsection{Rotation Augmentation for Object Detection}
An image is parameterized by $x$ and $y$ coordinates. Suppose the image contains an object with shape $S$. Let $S$ denote the \textbf{shape set} that describes all points in an object:
\begin{equation}
S = \{(x,y)\,\,|\,\,\forall\,\,(x,y) \in \textrm{object}\}.
\label{eq:S}
\end{equation}
In object detection, a bounding box is defined to be the tightest fitting axis-aligned box around a shape. Therefore a shape determines the coordinates of a bounding box, $\mathbf{b} = [x_{min},y_{min},x_{max},y_{max}]^T$. Each of the four edges of the bounding box intersects at least one element of the shape set. Let the operator $\mathcal{B}$ represent the perfect conversion of a shape to a bounding box: 
\begin{equation}
\mathbf{b}  \triangleq \mathcal{B}(S).   
\label{eq:bperfect1}
\end{equation}
\begin{figure*}
\centering
\includegraphics[width=\textwidth]{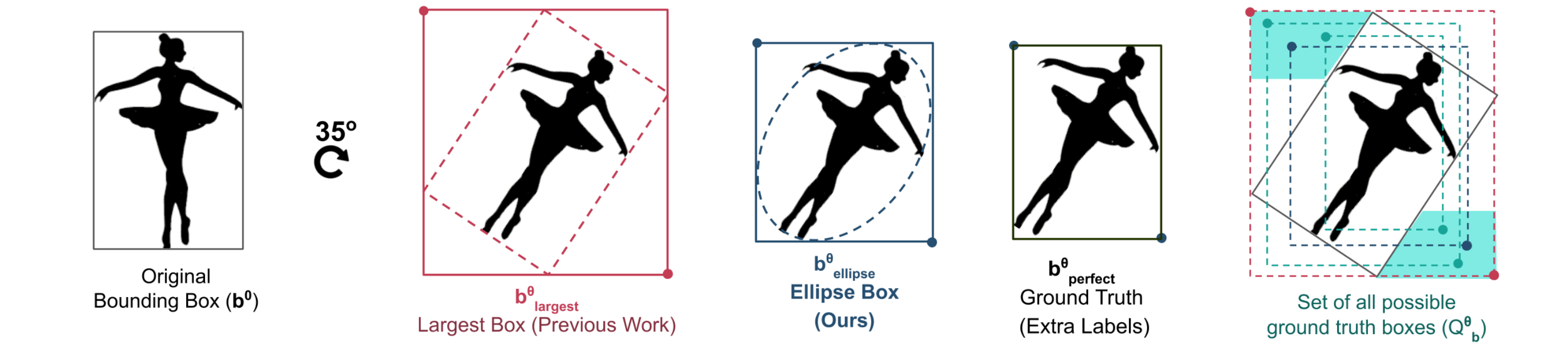} 
\caption{\textbf{Our Ellipse method leads to good initial training labels while the Largest Box overestimates the labels.} From left to right: (1) The original bounding box prior to rotation. (2) The oversized Largest Box estimate of the ground truth label post rotation. (3) The tighter Ellipse estimate (Section \ref{ss:boxlike}). (4) The actual ground truth which we get from segmentation shape labels. (5) The set of all possible ground truth boxes given a rotation and an initial box.}
\label{fig:ballet}
\label{fig:uncertainity}
\end{figure*}

The operator $\mathcal{B}$ extracts the bounding box $\mathbf{b}$ (tightest fitting axis aligned box) for $S$ by taking the minimum/maximum $(x, y)$ coordinates of the shape. As $S$ is not unique, the same bounding box can be generated by many shapes. For example, a square of side length $d$ and a circle of diameter $d$ are just two of many unique shapes that generate the same bounding box. More formally, let $V_\mathbf{b}$ denote the set of shapes that could possibly generate a bounding box $\mathbf{b}$, such that
\begin{equation}
    V_\mathbf{b} = \{ S_i\,\,|\,\, \mathcal{B}(S_i) = \mathbf{b}, \, \forall\,\,S_i \in \mc{P} \}.
    \label{eq:v}
\end{equation}

Where $\mc{P}$ is the dataset specific distribution of shapes. Let us consider the problem of rotation augmentation where an image and corresponding box label $\mathbf{b}$ is rotated by angle $\theta$. If the shape of the object is known, then we can rotate the original shape by angle $\theta$ using a rotation operator: $\mathcal{R}^{\theta}(S)$. In analogy to Equation~\ref{eq:bperfect1}, we can then  use the perfect method to obtain an axis-aligned bounding box for the rotated image as:

\begin{equation}
    \mathbf{b}^\theta_{\textrm{perfect}} \triangleq \mathcal{B}(\mc{R}^\theta({S}_{\textrm{perfect}})),
    \label{eq:bperfect2}
\end{equation}\

We call this method \textbf{perfect labels}, where ${S}_{\textrm{perfect}}$ is the actual shape of the object for a given bounding box $\mathbf{b}$. However, this requires shape labels, which are not available for object detection. In object detection, humans label boxes by implicitly segmenting the shape. Without knowing the shape labels, any shape $S \in V_\textbf{b}$ could be ${S}_{\textrm{perfect}}$, leading to many possible boxes $\mathbf{b^\theta}$. This paper seeks a method to estimate the rotated bounding box when we do not know the shape. We are only provided with the original bounding box $\mathbf{b}$, which we will hereafter write as $\mathbf{b}^0$ by making explicit that $\theta=0$. The problem statement follows.

\paragraph{Problem Statement:} Given only an input bounding box $\mathbf{b}^0$ and an angle $\theta$ by which the image should be rotated, find the axis aligned bounding box $\widehat{\mathbf{b}}^{\theta}$ that: (1) has high IoU with $\mathbf{b}^\theta_{\textrm{perfect}}$; and (2) improves model performance on rotated versions of vision datasets.

\subsection{Largest Valid Box Method}

Rotation augmentation without shape knowledge is not a new problem statement. The \emph{de facto} method in the object detection community for determining the bounding box post-rotation with no shape priors is the \textbf{largest box} method. The largest box method is \emph{extremely} prevalent (e.g.~\cite{montserrat2017training, liu2016novel, albumentations-team, chen2019mmdetection, abadi2016tensorflow, jung2018imgaug,casado2019clodsa, solt2019, kathuria_2018,kdnuggets, lozuwa_2019, saxena_2020, solawetz_2020, matlab}). Just like our proposed method, the largest box takes only the original bounding box $\mathbf{b}^0$ and $\theta$ as input. From Equation~\ref{eq:v} it is clear that several shapes could define $\mathbf{b}^0$. This creates an \emph{ambiguity problem}. The largest box method chooses the single largest of these possible shapes in area, $S_{\textrm{largest}}$. This shape is simply the box itself (Table \ref{tbshapey}). Treating this as the object shape, Equation~\ref{eq:bperfect2} can be adapted to obtain 
\begin{equation}
        \mathbf{b}^\theta_{\textrm{largest}} \triangleq \mathcal{B}(\mathcal{R}^\theta(S_{\textrm{largest}})).
        \label{eq:blargest}
\end{equation}

The benefit of this method is that it produces a box that is \emph{guaranteed} to contain the original object~\cite{zoph2019learning}, and it is easy to implement. The downside is that the method produces oversized boxes~\cite{fastai, zoph2019learning, exchange_1968, albumentations-team_q, open-mmlab_q, aleju}, and if used generously, hurts performance more than it helps (Table~\ref{table:final}). Surprisingly, to our best knowledge, including personal communication with practitioners and posts on internet forums, no alternatives have been adopted. We hope our method will change that.

\section{Proposed Solution}

\newcommand{\best}{\widehat{\mathbf{b}}^\theta}
\newcommand{\bestshape}{\widehat{S}}
\newcommand{\bperfect}{\mathbf{b}^\theta_{\textrm{perfect}}}
\newcommand{\bi}{\mathbf{b}^\theta_{j}}
\newcommand{\bk}{\mathbf{b}^\theta_{k}}

We now describe our solution to the problem: given $\mathbf{b^0}$ and desired rotation angle $\theta$, find $\best$. In a nutshell, our solution estimates a rotated bounded box by assuming the original shape is an ellipse (Table \ref{tbshapey}, Figure~\ref{fig:ballet}) and rotating accordingly (Section~\ref{ss:boxlike}). We then adapt the loss function to account for error in the labels (Section~\ref{ss:loss}).

\subsection{Ellipse Method}
\label{ss:boxlike}

In this section, we first derive the ellipse assumption from first principles by trying to find the shape that is most likely to have high overlap with potential ground truth boxes. Then we discuss the implementation and intuition of the ellipse method. Finally, we mention other novel methods we developed.

\subsubsection{Ellipse from Maximizing Expect IoU}
\label{ss:eiou}
We start with a simple assumption: the optimal method for determining a bounding box post-rotation augmentation should maximize label accuracy, which in the case of object detection is measured in IoU. 

We define $\best$ as the optimal rotated bounding box. We are provided the input angle $\theta$ and box $\mathbf{b}^0$. From Equation~\ref{eq:v}, this box could have been generated from any number of shapes: $V_{\mathbf{b}^0} = \{S_i\}_{i=1,2,\ldots,N}$. For each shape we can use the ``perfect method" from Equation~\ref{eq:bperfect2} to obtain a potential rotated bounding box. Since multiple shapes can lead to the same rotated box, we obtain $M \leq N$ possible bounding boxes, which we write as the set: 

\begin{align}
    Q^{\theta}_{\mathbf{b}^0} &= \textrm{unique}\{ \mathbf{b}^\theta_j = \mathcal{B}(\mathcal{R}^\theta(S_i)), \forall S_i \in V_{\mathbf{b}^0} \} \\
    &= \{\mathbf{b}^\theta_j\}_{j=1,\ldots,M}.
\label{eq:Q}
\end{align}

Hereafter, and without loss of generality, the paper will assume that $\mathbf{b}^0$ is the input allowing notation to be simplified: 

\begin{equation}
Q^\theta \triangleq Q^\theta_{\mathbf{b}^0}. 
\end{equation}

Now the task becomes to pick the ``best" of the $M$ possible bounding boxes in $Q^\theta$. Recall that the de facto solution is to choose the largest box in $Q^\theta$. This largest box is guaranteed to contain the object. However, optimizing for containment does not seem like a good choice to directly address the metric of AP because AP uses IoU to determine true positives, not containment. A more relevant goal for object detection is to select a box that maximizes:

\begin{equation}
    \best = \underset{\best \in Q^\theta}{\textrm{argmax}}  \,\, IoU(\best, \bperfect). 
\end{equation}
In which case $\best = \bperfect$. Of course, we are not given $\bperfect$. So for the moment, let us assume any of the bounding boxes in $Q^\theta$ has an equal chance of being the perfect box. Then, it would make sense to optimize over:

\begin{equation}
    \best = \underset{\best \in Q^\theta}{\textrm{argmax}}  \,\, Avg \{IoU(\best,\bi) \,\, | \,\, \forall \,\, \bi \in Q^\theta \}. 
    \label{eq:equal}
\end{equation}

Now, let us break the assumption that each candidate box in $Q^\theta$ is equally likely to be the perfect box. Indeed, we know that many shapes can produce the same box (since $M \leq N$), so certain boxes are more likely than others. For example, the only shape that can produce the largest box is the original box itself, whereas other rotated boxes can be generated by multiple object shapes in the dataset. Denote $p(\bi)$ as the probability that box $\bi = \bperfect$. Then Equation~\ref{eq:equal} can be reformulated as:

\begin{equation}
    \best = \underset{\best \in Q^\theta}{\textrm{argmax}} \sum_{i=1}^{M} p(\bi) IoU(\best,\bi).
\end{equation}

Readers may recognize this equation as being analogous to an expectation. We refer to this in the paper as the \textbf{Expected IoU}. The expected IoU is not directly tractable: we do not know $p(\bi)$ \emph{a priori}. However, if we can sample $K$ random shapes from a dataset distribution over shapes $S_k \sim \mathcal{P}$ where $\bk = \mc{B}(\mc{R}^\theta (S_k))$ we get the following optimization objective: 

\begin{equation}
    \best = \underset{\best \in Q^\theta}{\textrm{argmax}} \frac{1}{K}\sum^K_{k=1} IoU(\best,\mathbf{b}^\theta_{k}).
    \label{eq:bmaxiou}
\end{equation}

\begin{table}
\resizebox{\linewidth}{!}{
\begin{tabular}{l|l}
Method & Shape Definition \\
    \hline
        $S_{\textrm{largest}}$ & $\{(x_1, y_1), (x_2, y_1), (x_2, y_2), (x_1, y_2)\}$  \\ 
        $S_{\textrm{ellipse}} $ (Ours) & $ \Big\{ (x,y)\, \Big |\, \dfrac{(x-x_c)^2}{(\sfrac{b^0_W}{2})^2} + \dfrac{(y-y_c)^2}{(\sfrac{b^0_H}{2})^2} = 1 \Big\}$
    \end{tabular}
}
  \caption{\textbf{Our method can be compared with the Largest Box in the shape domain.} The implementation difference is one line of code.}
\label{tbshapey}
\end{table}

Since all object detection datasets do not have shape labels, we sample $\mc{P}$ by generating random shapes that touch each side of $b^0$ once. This way we are not dataset-specific. We analyze using COCO shapes in the supplement and show the performance is extremely similar. The above equation is fully differentiable, and so we can solve with gradient ascent. The problem here is that we would then have to solve this equation for every $\theta$ and every box $b$, and this is not practical. Therefore we generalize this further to a canonical shape.

\begin{figure}
    \centering
    \includegraphics[width=\linewidth]{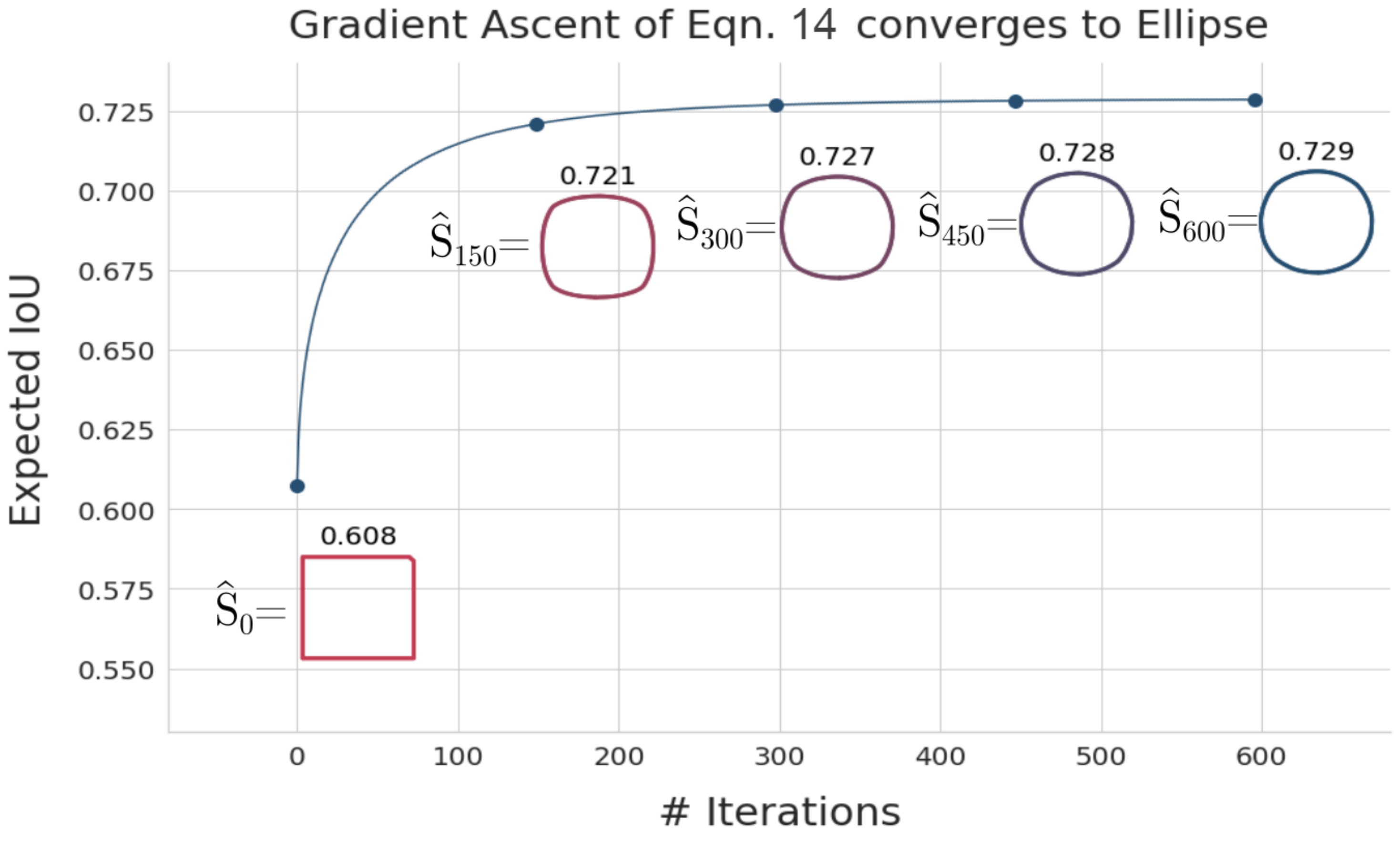}\label{fig:grad}
     
    \begin{tabular}{|l|c|c|c|}
    \hline
        \textbf{Label Method }& $S_{\textrm{largest}}$ & $S_{\textrm{ellipse}}$ (Ours) & $\bestshape$ (Ours) \\ \hline
        \textbf{Expected IoU} & \largestbox{60.8} & \textbf{72.9} & \textbf{72.9 }\\ \hline
    \end{tabular}
    
    \caption{\textbf{The optimal shape to maximize expected IoU with potential ground truth boxes converges to an ellipse.} Curve showing the progression of gradient ascent starting from the largest box and converging to an elliptical shape. The final converged expected IoU for $\bestshape$ matches that of the largest inscribed ellipse $S_{\textrm{ellipse}}$.}
    \label{fig:grad}
 \label{fig:iou}
 \end{figure} 

    
    \paragraph{Optimizing Equation~\ref{eq:bmaxiou} via a canonical shape:} Instead of solving Equation~\ref{eq:bmaxiou} for every possible combination of $b^0$ and $\theta$, we attempt to find a shape that is optimal across different input bounding boxes. This way, we could solve for some best shape $\bestshape \in V_{\mathbf{b^0}}$, and solve for $\best$ as follows:
    
\begin{equation}
    \best \triangleq \mathcal{B}( \mathcal{R}^\theta (\bestshape)),
\label{eq:Slikely}
\end{equation}

 To obtain the likely shape, we combine Equations~\ref{eq:bmaxiou} and \ref{eq:Slikely} to optimize the quantity:
 
\begin{equation}
    \bestshape = \underset{S \in \mc{V}_\mathbf{b^0}}{\textrm{argmax}} \, \sum_{\theta} \frac{1}{K} \sum_{k=1}^K \left[IoU(\mathcal{B}(\mathcal{R}^\theta(S)),\mathbf{b}^\theta_{k} )\right].
    \label{eq:brute}
\end{equation}

\begin{figure}
        \includegraphics[width=\linewidth]{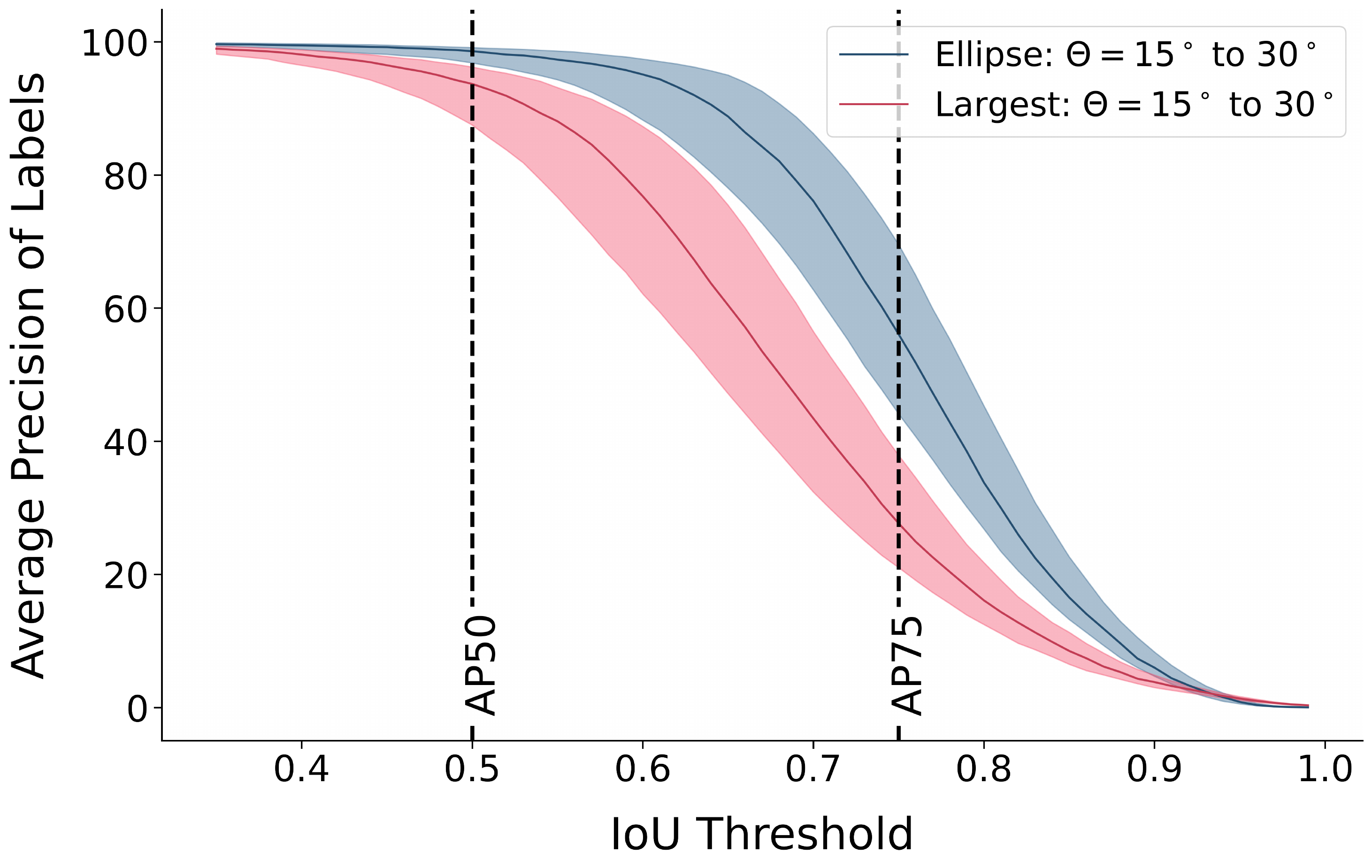}
         \resizebox{\linewidth}{!}{\begin{tabular}{|l|l|l|l|l||l|l|l|l|}
  \hline
       & \multicolumn{4}{c||}{\textbf{AP50}} & \multicolumn{4}{c|}{\textbf{AP75}} \\
        \cline{2-9}
          & $10^{\circ}$ & $20^{\circ}$ & $30^{\circ}$ & $40^{\circ}$ & $10^{\circ}$ & $20^{\circ}$ & $30^{\circ}$ & $40^{\circ}$\\ \hline
        Largest Box & \largestbox{98.2} & \largestbox{93.79} & \largestbox{86.31} & \largestbox{82.9} & \largestbox{59.2} & \largestbox{25.6} & \largestbox{19.9} & \largestbox{17.3} \\ \hline
           \setrow{\bfseries}  Ellipse (Ours) & \textbf{99.6} & \textbf{98.6} & \textbf{97.0} & \textbf{96.5} & \textbf{86.8} & \textbf{56.2} & \textbf{47.2} & \textbf{46.5}\\ \hline
    \end{tabular}}
    \caption{\textbf{Comparing Label AP (assuming uniform confidence) at different IOU thresholds for both methods at a $15^o-30^o$ rotation augmentation.} Ours is significantly better at AP50 and AP75.}
\label{fig:apcoco}
\end{figure}

Note that we now optimize over all rotation angles and aspect ratios simultaneously. This adds enough constraints to find a unique shape. The goal is to find \emph{the shape $\bestshape$ that produces an augmented bounding box that has high IoU with likely ground truth boxes}. Since $\mathcal{R}$ and $\mathcal{B}$ are differentiable operators, Equation~\ref{eq:brute} can be optimized through gradient ascent to solve for $\widehat{S}$. We provide details and pseudo-code and some analysis in the supplement. The stable solution found by gradient ascent is that of an elliptical shape. We show the progression of gradient ascent in Figure \ref{fig:grad} from the largest box shape to a circle. If we change the aspect ratio, it simply converges to the largest inscribed ellipse. Also in Figure \ref{fig:grad}, we show the Expected IoU for the Largest Box shape is much lower than the Ellipse, and in Figure \ref{fig:apcoco} we show that the resulting AP of the Ellipse labels is much better. The elliptical solution is similar to the optimized shape for various tested distributions of $\mathcal{P}$, including the random model described in the previous paragraph.
\subsubsection{The Ellipse Method}

\label{ss:ellipse}
 When we model the shape as an ellipse, we can find the estimated bounding box as:

\begin{equation}
    \best = \mathcal{B}( \mathcal{R}^\theta ( S_{\textrm{ellipse}} ) ),
\end{equation}
where $S_{\textrm{ellipse}}$ is the largest inscribed ellipse inside $\mathbf{b}^0$, expressed as:
\begin{equation}
S_\textrm{ellipse} = \Big\{ (x,y)\, \Big |\, \dfrac{(x-x_c)^2}{(\sfrac{b^0_W}{2})^2} + \dfrac{(y-y_c)^2}{(\sfrac{b^0_H}{2})^2} = 1 \Big\},
\end{equation}
where $(x_c,y_c)$ is the location of the center of $b^0$ and $b^0_W, b^0_H$ are the width and height of $b^0$ respectively. 

This equation is \textbf{fast, simple to implement, and high-performing on modern vision datasets}. The elliptical approximation can be implemented in the same line of code as the largest box method (cf. Appendix A), yet it greatly improves performance. We see in Figure \ref{fig:apcoco} the ellipse labels are far more accurate than the largest box labels. However, one disadvantage with the proposed elliptical box method is that the elliptical box can underestimate the object size or aspect ratio. This still causes some noise in the labels, especially at large rotations. We mitigate this by allowing the model to adapt labels at higher rotations based on priors from lower rotations in Section \ref{ss:loss}. 

\subsection{Other Methods}
\label{ss:other}
We do not limit our analysis to  the ellipse method. To perform a complete study we came up with an additional 4 methods. To conserve space, full details and results of these novel methods are available in the supplement Appendix B, we provide a quick summary for these methods here. 
\begin{itemize}
    \item \emph{Scaled Octagon:} We use an octagon with a scaling factor ($s$) to interpolate between the largest box shape and a diamond shape.
    \item \emph{Random Boxes:} We sample random valid boxes and use those as ground truth labels.
    \item \emph{RotIoU:} We select the label that has the maximum IoU with the rotated ground truth box rather than the expected axis-aligned ground truth box. 
    \item \emph{COCO Shape:} Rather than using random shapes for the optimization, we use the shapes from the COCO dataset. We keep results from this to the supplement since the performance between this and the ellipse method is negligible and we want this paper to be dataset independent and easy to implement.
\end{itemize}

\subsection{Rotation Uncertainty Loss}
\label{ss:loss}
\begin{figure}[t]
\begingroup\makeatletter\def\f@size{10}\check@mathfonts
$C(\theta) = \max\left(0.5, 1 + \frac{1 - \cos(4\theta)}{2\cos(4\delta) - 2}\right)$\endgroup

\centering
\includegraphics[width=0.9\linewidth]{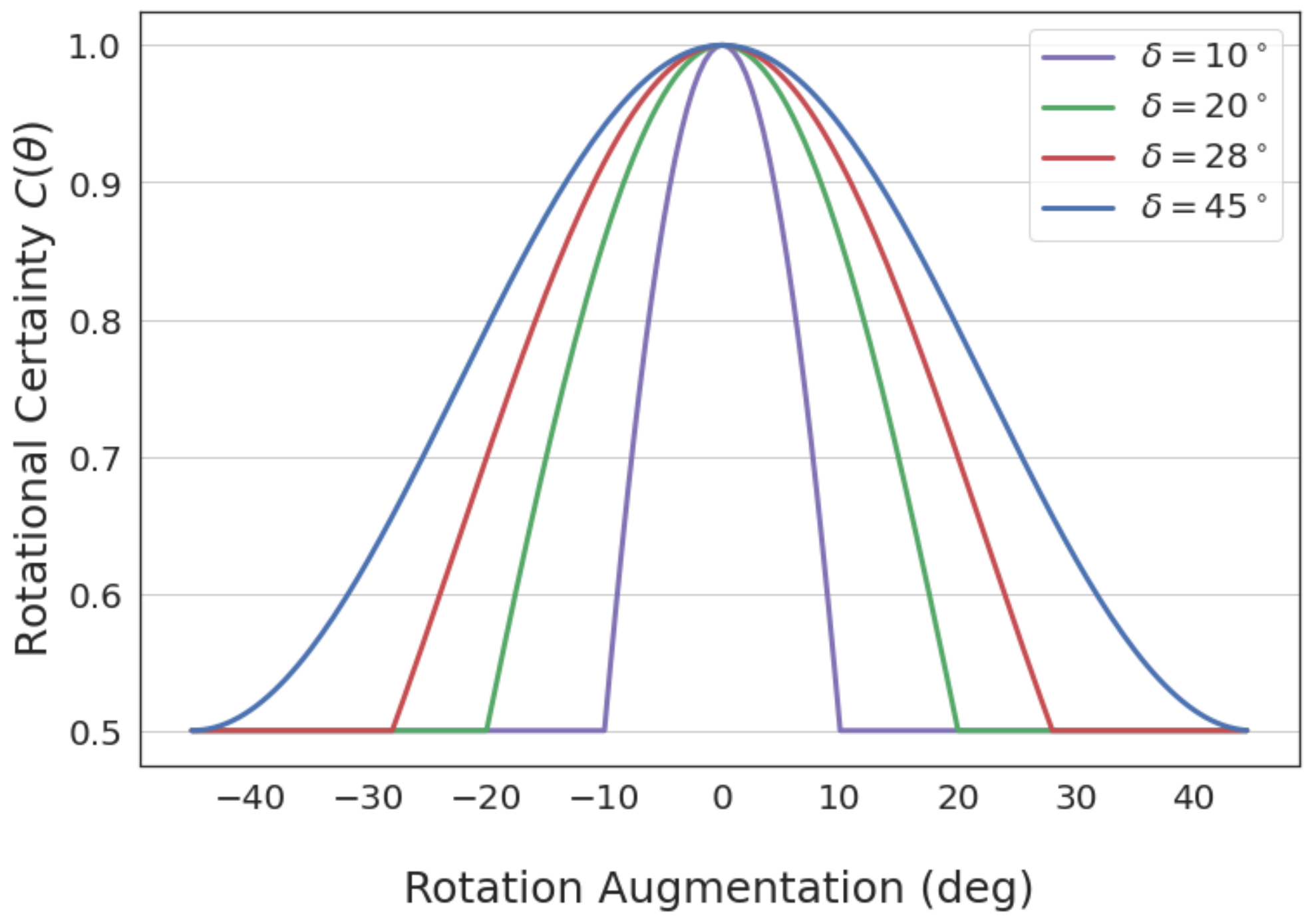}
  \caption{\textbf{Rotational certainty} used by RU Loss as a function of theta plotted for different hyper-parameters $\delta$ where $\delta$ is the angle at which $C(\theta) = 0.5$.}
  \label{fig:cos}
\end{figure}
\begin{figure}
\centering
\ffigbox{%
  \includegraphics[width=\linewidth]{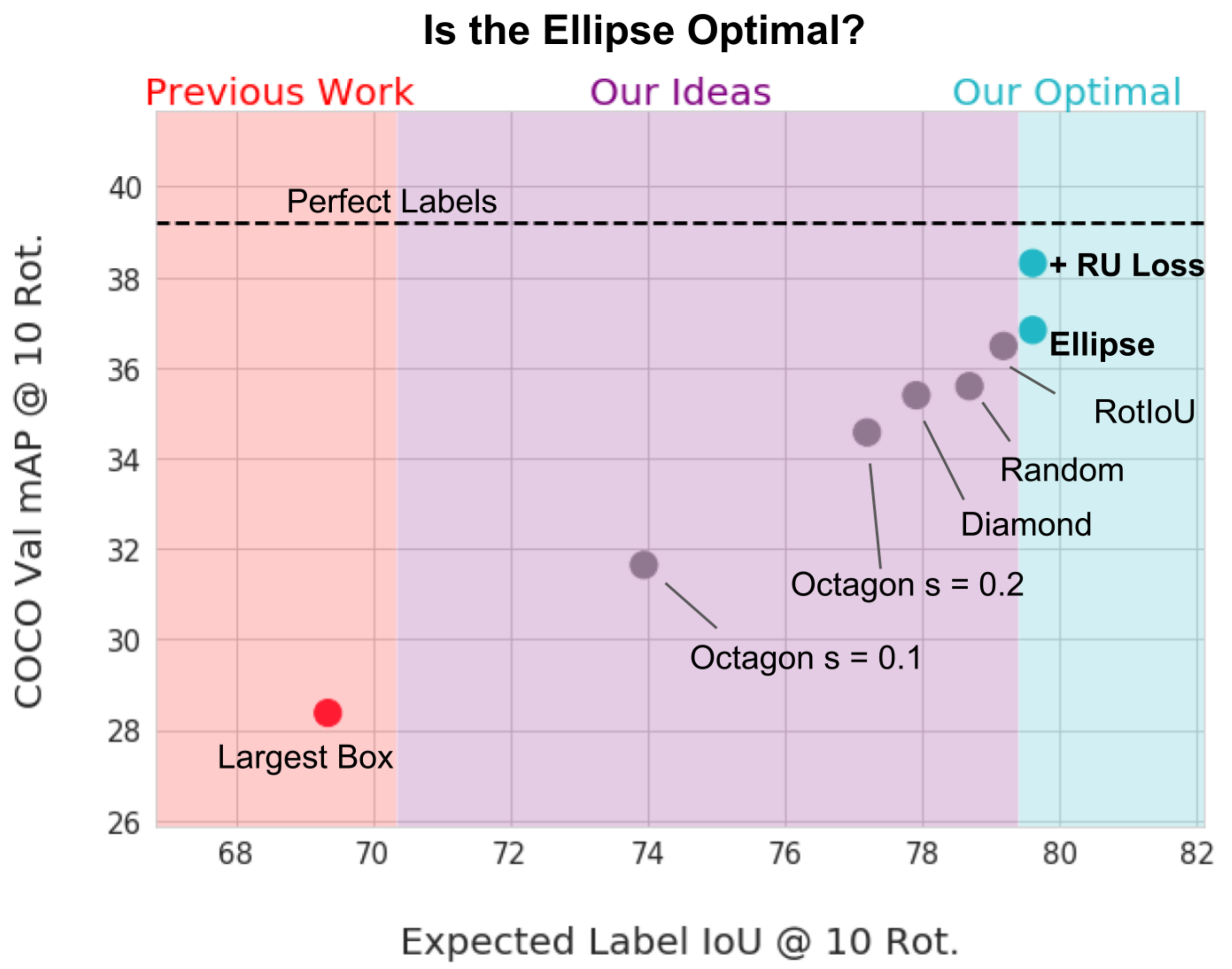} 
  }
  {\caption{\textbf{The Expected IoU of a rotation method is heavily correlated with performance.} Our final Ellipse is optimal for both.} \label{fig:eiouvsacc}}
   \capbtabbox{%
   \resizebox{\linewidth}{!}{
    \begin{tabular}{|l l l l l|}
    \hline
         \textbf{COCO val2017 Ablations} & \textbf{$0^{\circ}$} & \textbf{$10^{\circ}$} & \textbf{$20^{\circ}$} & \textbf{$30^{\circ}$}  \\ \hline
        \multicolumn{5}{|l|}{\small{\textit{(a) Previous method}}} \\ 
        Largest Box(e.g.~\cite{bochkovskiy2020yolov4, zoph2019learning, tan2019efficientnet})& 35.20 & 28.37 & 22.34 & 18.47 \\ \hline \hline
        \multicolumn{5}{|l|}{\small{\textit{(b) Our Rotation Label Methods (Section~\ref{ss:other})}}} \\ 
        Random & 37.39 & 35.59 & 32.22 & 28.33 \\ 
        Octagon $s=0.1$ & 35.82 & 31.64 & 27.16 & 23.54 \\ 
        Octagon $s=0.2$ & 36.52 & 34.57 & 31.65 & 28.15 \\ 
        Octagon $s=0.5$ (Diamond) & \textbf{38.36} & 35.39 & 28.76 & 22.92 \\
        RotIoU & \textbf{38.32} & 36.48 & 32.68 & 28.94 \\ 
    \hline
         \rowcolor{gray!30} \textbf{Ellipse} (Section~\ref{ss:ellipse})& \textbf{38.21} & \textbf{36.83} & \textbf{33.59} & \textbf{29.95} \\ 
        \hline\hline
        \multicolumn{5}{|l|}{\small{\textit{(c) With Our Loss (Section~\ref{ss:loss})}}} \\ 
        Ellipse + RU Loss $\delta=45^\circ$ & 38.54 & 37.45 & 34.56 & 31.26 \\ 
        Ellipse + RU Loss $\delta=30^\circ$ & 39.09 & 37.99 & 35.45 & 32.25 \\ 
        Ellipse + RU Loss $\delta=15^\circ$ & \textbf{39.14} & \textbf{38.19} & 35.78 & 32.50 \\ \hline
    \rowcolor{gray!30} \textbf{Ellipse + RU Loss  $\mathbf{\delta=10^\circ}$ (Final)}& \textbf{39.33} & \textbf{38.31} & \textbf{36.00} & \textbf{32.72} \\ \hline
    
    \end{tabular}}
    }
{\caption{\textbf{The AP at different test rotations on the COCO val2017 set for different methods.} (a) The previous method of largest box leads to the worst performance - every other idea we had was better. (b) The Ellipse is the best of all the label generation methods. (c) Our RU Loss with $\delta=10^\circ$ leads to the best AP across all rotations and therefore we use this as our final method. \emph{Note: We bold within 0.2 of best result.}} \label{table:abl}}
\vspace{-4 mm}
\end{figure}

\begin{figure*}[!htb]
\centering
{%
\ffigbox{
\includegraphics[width=\textwidth]{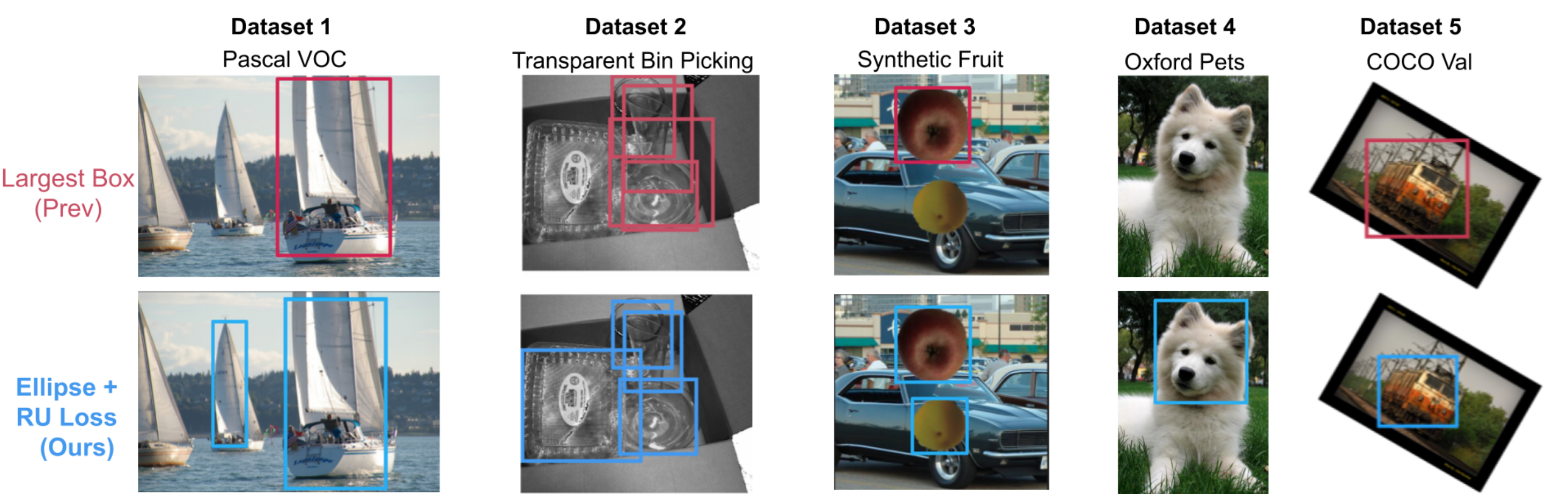}
}
  {\caption{\textbf{Example bounding box predictions for Largest Box model (top row) and Ellipse model (bottom row) for all 5 datasets}. We can see that ours produces tighter bounding boxes overall.}}

}%
 \capbtabbox{%
    \resizebox{\linewidth}{!}{\begin{tabular}{|l||l|l|l||l|l|l||l|l|l||l|l|l|}
    \hline
         Datasets (At $0^\circ$ Test Rotation) &   \multicolumn{3}{c||}{\textbf{Pascal VOC~\cite{everingham2010pascal}}} &   \multicolumn{3}{c||}{\textbf{Transparent Bin Picking~\cite{kalra_2020_cvpr}}} &   \multicolumn{3}{c||}{ \textbf{Synthetic Fruit Dataset} \cite{synthfruit}} &   \multicolumn{3}{c|}{\textbf{Oxford Pets} \cite{parkhi12a}}  \\ \hline
          Methods & \textbf{AP} &  \textbf{AP50} & \textbf{AP75} & \textbf{AP} & \textbf{AP50} & \textbf{AP75} & \textbf{AP} & \textbf{AP50} & \textbf{AP75} & \textbf{AP} & \textbf{AP50} & \textbf{AP75} \\ \hline
          
        No Rotation& 51.94 & 80.91 & 56.54 & 48.53 & 79.14 & 54.3 & 84.3 & 95.07 & 92.6 &80.70	&92.80	&88.76 \\ \hline \hline
        
        Largest Box(e.g.~\cite{bochkovskiy2020yolov4, zoph2019learning, tan2019efficientnet}) & 50.23 & 81.31 & 54.3 & 37.49 & 79.09 & 28.45 & 83.47 & 95.05 & 92.24 &79.54	&94.20	& 90.03 \\ 
        		
          (relative improvement)  & \textcolor{red}{-3.29\%} & 0.49\% & \textcolor{red}{-3.96\%} & \textcolor{red}{-22.7\%} & \textcolor{red}{-0.06\%} & \textcolor{red}{-47.6\%} & \textcolor{red}{-0.98\%} & \textcolor{red}{-0.02\%} &
          \textcolor{red}{-0.39\%} & \textcolor{red}{-1.43\%} & {1.56\%} & {1.43\%} \\ \hline \hline
       \rowcolor{gray!30} \textbf{Ellipse + RU Loss (Ours)}   & \textbf{52.89} & \textbf{81.57} & \textbf{57.97} & \textbf{50.36} &  \textbf{81.78} & \textbf{56.76} & \textbf{84.83} & \textbf{95.83} & \textbf{93.17} & \textbf{81.28}	& \textbf{94.37}	& \textbf{91.09} \\ 
\rowcolor{gray!30}(relative improvement) & \textbf{1.84\%} & \textbf{0.82\%} & \textbf{2.53\%} & \textbf{3.78\%} & \textbf{3.35\% }& \textbf{4.53\%} & \textbf{1.05\%} &\textbf{ 0.80\%} & \textbf{0.62\%} & \textbf{0.72\%} & \textbf{1.69\%} & \textbf{2.63\%} \\ \hline \hline
    \end{tabular}
    }
    }{%
    \caption{\textbf{Across four separate datasets we show that our method of rotation augmentation leads to an improved performance where the previous method hurts performance.} Especially in the case of transparent object bin picking where the largest box is almost 50\% worse and ours is 4.5\% better AP75. \label{tbl:datasets}}
}
\end{figure*}

 As shown in Figure \ref{fig:grad} the expected IoU with random shapes is 72.9. This means attaining good performance at the higher APs, like AP75, will be very difficult using just these labels. To tackle this problem, we create a custom loss function that adapts the regression loss to account for the uncertainty of the rotation. The idea is simple - if we are uncertain of the label, we turn off the regression loss if the model is \emph{close enough}. The labels are more uncertain as the rotation approaches $45^\circ, 135^\circ, 225^\circ, 315^\circ$. and perfectly certain at $0^\circ, 90^\circ, 180^\circ \text{ and } 270^\circ.$ We formalize on the concept of certainty (Figure~\ref{fig:cos}) as a function of $\theta$:

\begin{equation} 
\alpha = 2\cos(4\delta),
\end{equation}
\begin{equation}
C(\theta) = 1 + \frac{1}{\alpha-2}(1 - \cos(4\theta)).
\end{equation}

This function maps a rotation $\theta$ to an IoU threshold $C(\theta)$. We use this IoU threshold to serve as an indicator for applying regression loss. If the predicted box is greater than $max(0.5, C(\theta))$, it uses the regression loss, otherwise, it does not and assumes the model's prediction is correct. We parameterize $C$ with $\delta$. $\delta$ is the angle at which $C(\theta) = 0.5$. We visualize $C$ in \Cref{fig:cos}. We bound it by 0.5 since that is the threshold for anchor-matching in standard object detection architectures \cite{lin2017focal}. 

This function allows the model to take the priors it learns at the confident rotations and apply them to the higher rotations, preventing it from overfitting to poor labels. We show in Table \ref{table:abl}.

\section{Results}

\subsection{Setup}
Our hardware setup contains only a single P100 GPU for training, and all our code is implemented in Detectron2~\cite{wu2019detectron2} with Pytorch~\cite{ paszke2019pytorch}. We use the default training pipeline for both Faster-RCNN~\cite{ren2015faster} and RetinaNet~\cite{lin2017focal}. We conduct most of our experiments on the standard COCO benchmark since it contains a variety of objects with many different shapes - making it a challenging test set.

\textbf{Training}
Since we have only a single GPU, we can only fit a batch size of 3. To account for this, we increase the training time by around 5x from the default configurations. This allows us to match online available pre-trained baselines for RetinaNet~\cite{lin2017focal} and Faster-RCNN~\cite{ren2015faster}. Since most datasets are right-side-up images, we train with a normal distribution with a mean of 0 and a standard deviation of 15 degrees for all experiments. Since this paper aims to find the optimal rotation augmentation method, not the strategy for applying rotation augmentation, we do not try other combinations. This may be left for future work. 
\begin{figure*}
\begin{center}
\includegraphics[width=0.9\textwidth]{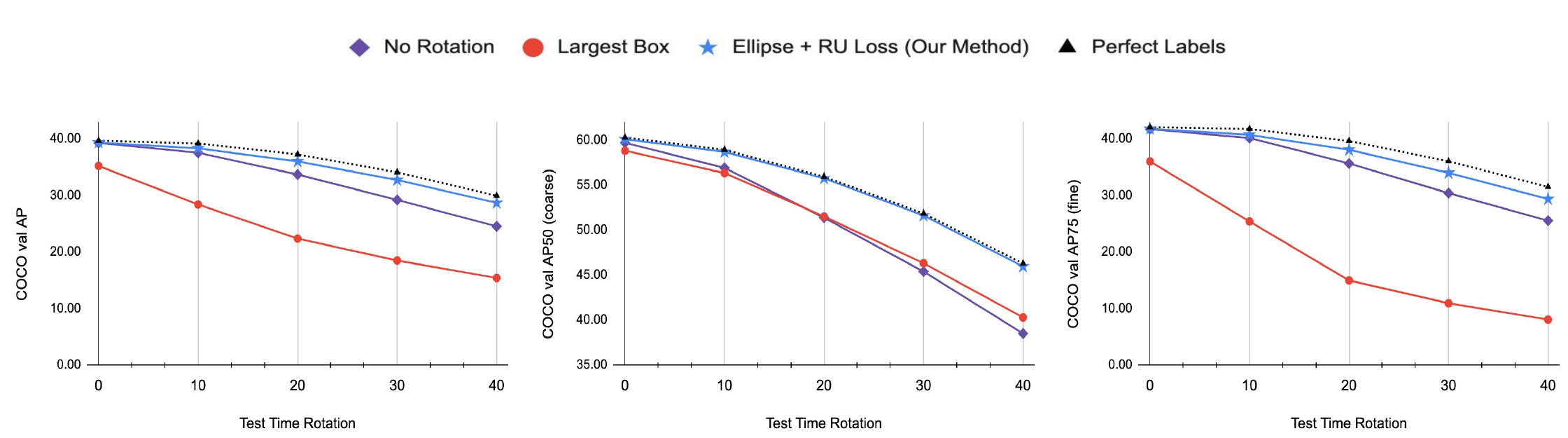}
\end{center}
\label{fig:final}
\resizebox{0.9\textwidth}{!}{
    \begin{tabular}{|l|l l l l||l l l l||l l l l|}
    \hline
        \multirow{2}{*}{\textbf{COCO val2017 Results}} &  \multicolumn{4}{c||}{\textbf{AP}}  & \multicolumn{4}{c||}{\textbf{AP50}}  & \multicolumn{4}{c|}{\textbf{AP75}}   \\ \cline{2-13}
         &  $\boldsymbol{0^{\circ}}$ & $\boldsymbol{{10}^{\circ}}$ & $\boldsymbol{{20}^{\circ}}$ &  $\boldsymbol{{30}^{\circ}}$ &    $\boldsymbol{0^{\circ}}$ & $\boldsymbol{{10}^{\circ}}$ & $\boldsymbol{{20}^{\circ}}$ &  $\boldsymbol{{30}^{\circ}}$ &   $\boldsymbol{0^{\circ}}$ & $\boldsymbol{{10}^{\circ}}$ & $\boldsymbol{{20}^{\circ}}$ &  $\boldsymbol{{30}^{\circ}}$   \\ \hline
        No Rotation & 39.26 & 37.54 & 33.68 & 29.19 & 59.68 & 56.88 & 51.35 & 45.37 &  41.69 & 40.14 & 35.63 & 30.39 \\ \hline
        
        Largest Box (e.g.~\cite{bochkovskiy2020yolov4, zoph2019learning, tan2019efficientnet}) & 35.20 & 28.37 & 22.34 & 18.47 & 58.79 & 56.31 & 51.49 & 46.30 &  36.00 & 25.37 & 14.95 & 10.91 \\ 
         (relative improvement) & \textcolor{red}{-10.3\%} & \textcolor{red}{-24.4\%} & \textcolor{red}{-33.7\%} & \textcolor{red}{-36.7\%} &  \textcolor{red}{-1.48\%} & \textcolor{red}{-1.01\%} & \textcolor{black}{0.27\%} & \textcolor{black}{2.04\%} & \textcolor{red}{-13.6\%} & \textcolor{red}{-36.8\%} & \textcolor{red}{-58.1\%} & \textcolor{red}{-64.1\%}  \\ \hline  \hline
         
        \rowcolor{gray!30}  \textbf{Ellipse + RU Loss (Ours)}  & \textbf{39.33} & \textbf{38.31 }& \textbf{36.00} & \textbf{32.72} & \textbf{60.08} & \textbf{58.66} & \textbf{55.73} & \textbf{51.60} & \textbf{41.74} & \textbf{40.71} & \textbf{38.05} & \textbf{33.97}  \\ 
        \rowcolor{gray!30}  \textbf{(relative improvement)} & \textbf{0.17\%} & \textbf{2.05\%} & \textbf{6.88\%} & \textbf{12.1\%}  & \textbf{0.67\%} & \textbf{3.12\%} & \textbf{8.54\%} & \textbf{13.7\%} &  \textbf{0.13\%} &\textbf{ 1.42\%} & \textbf{6.79\%} & \textbf{11.8\%} \\ \hline  \hline
        Perfect Labels & 39.66 & 39.17 & 37.24 & 34.08 & 60.28 & 58.89 & 55.91 & 51.83 & 42.05 & 41.73 & 39.61 & 36.02  \\ 
                (relative improvement) & 1.02\% & 4.36\% & 10.6\% & 16.7\% & 1.00\% & 3.53\% & 8.88\% & 14.2\% & 0.88\% & 3.94\% & 11.2\% & 18.5\%  \\ \hline  
    \end{tabular}
    }
     \caption{\textbf{Our method of Ellipse + RU Loss performs close to perfect labels for AP50} and performs better than No Rotation and Largest Box for AP and AP75 - demonstrating the first reliable rotation augmentation without shape labels. \label{table:final}}
     
\end{figure*}

\textbf{Testing:} For all datasets except COCO we do not have complete segmentation labels, so we only test on the standard test set ($0^\circ$). For COCO we generate our test set by taking the COCO val2017 set and rotating it from $0^\circ$-$40^\circ$ to simulate out-of-distribution rotations. We then bucket these rotations in intervals of 10 and evaluate using segmentation labels to generate ground truth. We leave COCO results for Faster-RCNN to supplement because they are similar to the results for RetinaNet shown below.

\subsection{Ablation Studies}
In Figure \ref{fig:eiouvsacc} and the accompanying table, we conduct a thorough ablation study on both the method for choosing the label and the impact of the RU Loss function. 

\paragraph{Justifying EIoU Optimization:} In Section \ref{ss:eiou} we assumed that the optimal method for label rotation should maximize label accuracy, which we approximated as Expected IoU (Eq. \ref{eq:brute}). In Figure \ref{fig:eiouvsacc} we demonstrate a strong correlation between Expected IoU and performance on COCO at $10^\circ$ across all methods, proving the effectiveness of our first principles derivation. We see similar correlations at other angles as well.

\paragraph{Justifying the Ellipse:} In Section \ref{ss:boxlike} we introduced many potential methods for rotating a box label. In the ablation Table \ref{table:abl}b we show that the Ellipse leads to the best performance across all rotations except $0^\circ$ where it is within a small noise tolerance. It is also important to note that all methods we tried perform significantly better than the Largest Box - showing the importance of fixing this issue.

\paragraph{RU Loss Ablation:} In Section \ref{ss:loss} we introduce a hyper-parameter $\delta$ in our final method. We ablate over that in Table \ref{table:abl} and demonstrate that $10^\circ$ is optimal. We found this to be true on COCO, however, on simpler datasets we use larger values of $\delta$ for optimal performance.

\subsection{Overall Performance}
Our best performing method consists of using both Ellipse-based label rotation and RU Loss. In this section, we show it leads to much better performance across multiple datasets and approximates segmentation-based rotation augmentations on COCO.
\subsubsection{Object Detection Datasets}

In Table \ref{tbl:datasets}, we provide four datasets where our method of rotation augmentation improves performance while the previous one (Largest Box) hurts performance. We notice this to be especially bad at higher APs, such as AP75. The gap is also larger in complex datasets such as transparent object bin picking where the largest box reduces performance by almost 50\% and ours increases it by 4.5\%.
\subsubsection{Generalizing to new Rotation Angles}
 Our method significantly outperforms the original largest box method and also outperforms not using rotation for AP, AP50, and AP75 across all new angles from $[0^o-30^o]$ on COCO in Figure \ref{fig:first} and Figure \ref{table:final}. In the case of AP50, we show very similar improvements compared to using segmentation-based labels. This is a huge improvement since the largest box method hurts rotation performance.

\section{Conclusion}
 The widespread Largest Box method (e.g.~\cite{xi2018sr, bochkovskiy2020yolov4, zoph2019learning, tan2019efficientnet, montserrat2017training, liu2016novel}) is based on the folk wisdom of maximizing overlap. Instead, we show that by maximizing Expected IoU and accounting for label certainty in the loss, we can \textbf{completely match} the performance of perfect ``segmentation-based" labels at AP50 while also achieving \textbf{significant gains} for AP and AP75. These results represent a step toward achieving rotation invariance for object detection models, while adding only a few lines of complexity to object detection codebases. 

 \paragraph{Acknowledgements:} We thank Yuri Boykov, Tomas Gerlich, Abhijit Ghosh, Olga Veksler and Kartik Venkataraman for their helpful discussions and edits to improve the paper.

\bibliographystyle{ieee_fullname}
\bibliography{egbib}

\begin{thebibliography}{10}\itemsep=-1pt

\bibitem{abadi2016tensorflow}
Mart{\'\i}n Abadi, Paul Barham, Jianmin Chen, Zhifeng Chen, Andy Davis, Jeffrey
  Dean, Matthieu Devin, Sanjay Ghemawat, Geoffrey Irving, Michael Isard, et~al.
\newblock Tensorflow: A system for large-scale machine learning.
\newblock In {\em 12th $\{$USENIX$\}$ symposium on operating systems design and
  implementation ($\{$OSDI$\}$ 16)}, pages 265--283, 2016.

\bibitem{albumentations-team_q}
Albumentations-Team.
\newblock Why does the bounding box become so loose after rotation data aug? is
  there a better way? · issue 746 · albumentations-team/albumentations.
\newblock
  \url{https://github.com/albumentations-team/albumentations/issues/746}.

\bibitem{aleju}
Aleju.
\newblock Make the bounding box more tighten for the object after image
  rotation · issue 90 · aleju/imgaug.
\newblock \url{https://github.com/aleju/imgaug/issues/90}.

\bibitem{bochkovskiy2020yolov4}
Alexey Bochkovskiy, Chien-Yao Wang, and Hong-Yuan~Mark Liao.
\newblock Yolov4: Optimal speed and accuracy of object detection.
\newblock {\em arXiv preprint arXiv:2004.10934}, 2020.

\bibitem{albumentations-team}
Alexander Buslaev, Vladimir~I Iglovikov, Eugene Khvedchenya, Alex Parinov,
  Mikhail Druzhinin, and Alexandr~A Kalinin.
\newblock Albumentations: fast and flexible image augmentations.
\newblock {\em Information}, 11(2):125, 2020.

\bibitem{casado2019clodsa}
{\'A}ngela Casado-Garc{\'\i}a, C{\'e}sar Dom{\'\i}nguez, Manuel
  Garc{\'\i}a-Dom{\'\i}nguez, J{\'o}nathan Heras, Adri{\'a}n In{\'e}s, Eloy
  Mata, and Vico Pascual.
\newblock Clodsa: a tool for augmentation in classification, localization,
  detection, semantic segmentation and instance segmentation tasks.
\newblock {\em BMC bioinformatics}, 20(1):323, 2019.

\bibitem{chen2019mmdetection}
Kai Chen, Jiaqi Wang, Jiangmiao Pang, Yuhang Cao, Yu Xiong, Xiaoxiao Li,
  Shuyang Sun, Wansen Feng, Ziwei Liu, Jiarui Xu, et~al.
\newblock Mmdetection: Open mmlab detection toolbox and benchmark.
\newblock {\em arXiv preprint arXiv:1906.07155}, 2019.

\bibitem{cheng2018learning}
Gong Cheng, Junwei Han, Peicheng Zhou, and Dong Xu.
\newblock Learning rotation-invariant and fisher discriminative convolutional
  neural networks for object detection.
\newblock {\em IEEE Transactions on Image Processing}, 28(1):265--278, 2018.

\bibitem{cheng2016learning}
Gong Cheng, Peicheng Zhou, and Junwei Han.
\newblock Learning rotation-invariant convolutional neural networks for object
  detection in vhr optical remote sensing images.
\newblock {\em IEEE Transactions on Geoscience and Remote Sensing},
  54(12):7405--7415, 2016.

\bibitem{cheng2016rifd}
Gong Cheng, Peicheng Zhou, and Junwei Han.
\newblock Rifd-cnn: Rotation-invariant and fisher discriminative convolutional
  neural networks for object detection.
\newblock In {\em Proceedings of the IEEE conference on computer vision and
  pattern recognition}, pages 2884--2893, 2016.

\bibitem{devries2017improved}
Terrance DeVries and Graham~W Taylor.
\newblock Improved regularization of convolutional neural networks with cutout.
\newblock {\em arXiv preprint arXiv:1708.04552}, 2017.

\bibitem{Ding_2019_CVPR}
Jian Ding, Nan Xue, Yang Long, Gui-Song Xia, and Qikai Lu.
\newblock Learning roi transformer for oriented object detection in aerial
  images.
\newblock In {\em The IEEE Conference on Computer Vision and Pattern
  Recognition (CVPR)}, June 2019.

\bibitem{synthfruit}
Brad Dwyer.
\newblock How to create a synthetic dataset for computer vision.
\newblock
  \url{https://blog.roboflow.com/how-to-create-a-synthetic-dataset-for-computer-vision/},
  Aug 2020.

\bibitem{everingham2010pascal}
Mark Everingham, Luc Van~Gool, Christopher~KI Williams, John Winn, and Andrew
  Zisserman.
\newblock The pascal visual object classes (voc) challenge.
\newblock {\em International journal of computer vision}, 88(2):303--338, 2010.

\bibitem{Goodfellow-et-al-2016}
Ian Goodfellow, Yoshua Bengio, and Aaron Courville.
\newblock {\em Deep Learning}.
\newblock MIT Press, 2016.
\newblock \url{http://www.deeplearningbook.org}.

\bibitem{fastai}
Jeremy Howard.
\newblock Lesson 9: Deep learning part 2 2018 - multi-object detection - fast
  ai.
\newblock \url{https://youtu.be/0frKXR-2PBY?t=554}.

\bibitem{jung2018imgaug}
Alexander~B Jung, K Wada, J Crall, S Tanaka, J Graving, S Yadav, J Banerjee, G
  Vecsei, A Kraft, J Borovec, et~al.
\newblock imgaug, 2018.

\bibitem{kalra_2020_cvpr}
Agastya Kalra, Vage Taamazyan1, Supreeth Krishna~Raol, Kartik Venkataraman,
  Ramesh Raskar, and Achuta Kadambi.
\newblock Deep polarization cues for transparent object segmentation.
\newblock In {\em The IEEE Conference on Computer Vision and Pattern
  Recognition (CVPR)}, June 2020.

\bibitem{kdnuggets}
Ayoosh Kathuria.
\newblock Data augmentation for bounding boxes: Rethinking image transforms for
  object detection.
\newblock
  \url{https://www.kdnuggets.com/2018/09/data-augmentation-bounding-boxes-image-transforms.html}.

\bibitem{kathuria_2018}
Ayoosh Kathuria.
\newblock Data augmentation for object detection: How to rotate bounding boxes.
\newblock
  \url{https://blog.paperspace.com/data-augmentation-for-object-detection-rotation-and-shearing/},
  Sep 2018.

\bibitem{lin2017focal}
Tsung-Yi Lin, Priya Goyal, Ross Girshick, Kaiming He, and Piotr Doll{\'a}r.
\newblock Focal loss for dense object detection.
\newblock In {\em Proceedings of the IEEE international conference on computer
  vision}, pages 2980--2988, 2017.

\bibitem{lin2014microsoft}
Tsung-Yi Lin, Michael Maire, Serge Belongie, James Hays, Pietro Perona, Deva
  Ramanan, Piotr Doll{\'a}r, and C~Lawrence Zitnick.
\newblock Microsoft coco: Common objects in context.
\newblock In {\em European conference on computer vision}, pages 740--755.
  Springer, 2014.

\bibitem{liu2018rotation}
Baozhen Liu, Hang Wu, Weihua Su, Wenchang Zhang, and Jinggong Sun.
\newblock Rotation-invariant object detection using sector-ring hog and boosted
  random ferns.
\newblock {\em The Visual Computer}, 34(5):707--719, 2018.

\bibitem{liu2014rotation}
Kun Liu, Henrik Skibbe, Thorsten Schmidt, Thomas Blein, Klaus Palme, Thomas
  Brox, and Olaf Ronneberger.
\newblock Rotation-invariant hog descriptors using fourier analysis in polar
  and spherical coordinates.
\newblock {\em International Journal of Computer Vision}, 106(3):342--364,
  2014.

\bibitem{liu2017learning}
Lei Liu, Zongxu Pan, and Bin Lei.
\newblock Learning a rotation invariant detector with rotatable bounding box.
\newblock {\em arXiv preprint arXiv:1711.09405}, 2017.

\bibitem{liu2016novel}
Yang Liu, Lei Huang, Xianglong Liu, and Bo Lang.
\newblock A novel rotation adaptive object detection method based on pair hough
  model.
\newblock {\em Neurocomputing}, 194:246--259, 2016.

\bibitem{lozuwa_2019}
Rodrigo Lozuwa.
\newblock lozuwa/impy.
\newblock \url{https://github.com/lozuwa/impy}, Mar 2019.

\bibitem{montserrat2017training}
Daniel~Mas Montserrat, Qian Lin, Jan Allebach, and Edward~J Delp.
\newblock Training object detection and recognition cnn models using data
  augmentation.
\newblock {\em Electronic Imaging}, 2017(10):27--36, 2017.

\bibitem{open-mmlab_q}
Open-Mmlab.
\newblock Why does the bounding box become so loose after rotation data aug? is
  there a better way · issue 4070 · open-mmlab/mmdetection.
\newblock \url{https://github.com/open-mmlab/mmdetection/issues/4070}.

\bibitem{parkhi12a}
Omkar~M. Parkhi, Andrea Vedaldi, Andrew Zisserman, and C.~V. Jawahar.
\newblock Cats and dogs.
\newblock In {\em IEEE Conference on Computer Vision and Pattern Recognition},
  2012.

\bibitem{paszke2019pytorch}
Adam Paszke, Sam Gross, Francisco Massa, Adam Lerer, James Bradbury, Gregory
  Chanan, Trevor Killeen, Zeming Lin, Natalia Gimelshein, Luca Antiga, et~al.
\newblock Pytorch: An imperative style, high-performance deep learning library.
\newblock In {\em Advances in neural information processing systems}, pages
  8026--8037, 2019.

\bibitem{ren2015faster}
Shaoqing Ren, Kaiming He, Ross Girshick, and Jian Sun.
\newblock Faster r-cnn: Towards real-time object detection with region proposal
  networks.
\newblock In {\em Advances in neural information processing systems}, pages
  91--99, 2015.

\bibitem{saxena_2020}
Pranjal Saxena.
\newblock Data augmentation for custom object detection: Yolo.
\newblock
  \url{https://medium.com/predict/data-augmentation-for-custom-object-detection-15674966e0c8},
  Sep 2020.

\bibitem{schmidt2012learning}
Uwe Schmidt and Stefan Roth.
\newblock Learning rotation-aware features: From invariant priors to
  equivariant descriptors.
\newblock In {\em 2012 IEEE Conference on Computer Vision and Pattern
  Recognition}, pages 2050--2057. IEEE, 2012.

\bibitem{exchange_1968}
Dennis Soemers.
\newblock How data augmentation like rotation affects the quality of detection?
\newblock
  \url{https://ai.stackexchange.com/questions/9935/how-data-augmentation-like-rotation-affects-the-quality-of-detection},
  Mar 2018.

\bibitem{solawetz_2020}
Jacob Solawetz.
\newblock Why and how to implement random rotate data augmentation.
\newblock
  \url{https://blog.roboflow.com/why-and-how-to-implement-random-rotate-data-augmentation/},
  Aug 2020.

\bibitem{tan2019efficientnet}
Mingxing Tan and Quoc~V Le.
\newblock Efficientnet: Rethinking model scaling for convolutional neural
  networks.
\newblock {\em arXiv preprint arXiv:1905.11946}, 2019.

\bibitem{taylor2018improving}
Luke Taylor and Geoff Nitschke.
\newblock Improving deep learning with generic data augmentation.
\newblock In {\em 2018 IEEE Symposium Series on Computational Intelligence
  (SSCI)}, pages 1542--1547. IEEE, 2018.

\bibitem{matlab}
Matlab Team.
\newblock Augment bounding boxes for object detection.
\newblock
  \url{https://www.mathworks.com/help/deeplearning/ug/bounding-box-augmentation-using-computer-vision-toolbox.html}.

\bibitem{solt2019}
Aleksei Tiulpin.
\newblock Solt: Streaming over lightweight transformations, July 2019.

\bibitem{wang2015ordinal}
Guoli Wang, Bin Fan, and Chunhong Pan.
\newblock Ordinal pyramid pooling for rotation invariant object recognition.
\newblock In {\em 2015 IEEE International Conference on Acoustics, Speech and
  Signal Processing (ICASSP)}, pages 1349--1353. IEEE, 2015.

\bibitem{wu2019detectron2}
Yuxin Wu, Alexander Kirillov, Francisco Massa, Wan-Yen Lo, and Ross Girshick.
\newblock Detectron2, 2019.

\bibitem{xi2018sr}
Yue Xi, Jiangbin Zheng, Xiuxiu Li, Xinying Xu, Jinchang Ren, and Gang Xie.
\newblock Sr-pod: sample rotation based on principal-axis orientation
  distribution for data augmentation in deep object detection.
\newblock {\em Cognitive Systems Research}, 52:144--154, 2018.

\bibitem{yun2019cutmix}
Sangdoo Yun, Dongyoon Han, Seong~Joon Oh, Sanghyuk Chun, Junsuk Choe, and
  Youngjoon Yoo.
\newblock Cutmix: Regularization strategy to train strong classifiers with
  localizable features.
\newblock In {\em Proceedings of the IEEE International Conference on Computer
  Vision}, pages 6023--6032, 2019.

\bibitem{zoph2019learning}
Barret Zoph, Ekin~D Cubuk, Golnaz Ghiasi, Tsung-Yi Lin, Jonathon Shlens, and
  Quoc~V Le.
\newblock Learning data augmentation strategies for object detection.
\newblock {\em arXiv preprint arXiv:1906.11172}, 2019.

\end{thebibliography}

\end{document}